\documentclass[letterpaper, 10 pt, journal, twoside]{IEEEtran} 
\IEEEoverridecommandlockouts

\usepackage{cite}
\usepackage{amsmath,amssymb,amsfonts}
\usepackage{bm}
\usepackage{graphicx}
\usepackage{textcomp}
\usepackage{xcolor}
\usepackage{booktabs}
\usepackage{multirow}
\usepackage[caption=false,font=footnotesize]{subfig}
\usepackage{url}
\usepackage{gensymb}
\usepackage{threeparttable}
\usepackage{balance}

\usepackage{enumitem}
\begin{document}

\title{BEV-Diffusion: Camera-Only Traversability Navigation  with BEV Memory and Efficient Cross-Robot Adaptation}

\author{
  \IEEEauthorblockN{Iana Zhura, Timofei Kozlov, Dmitrii Plotnikov, Ivan Valuev, \\ Miguel Altamirano Cabrera,  Dzmitry Tsetserukou}
  \thanks{%
    The authors are with the Intelligent Space Robotics Laboratory\\
    Skolkovo Institute of Science and Technology
    Moscow, Russia\\
    \tt\footnotesize{\{i.zhura, timofei.kozlov, dmitrii.plotnikov, ivan.valuev, m.altamirano d.tsetserukou\}@skoltech.ru}}
}

\maketitle
\thispagestyle{empty}

\begin{abstract}
Robot navigation requires both an understanding of the scene and an efficient adaptation to new robots, yet current methods rely on expensive LiDAR sensors, long mapping and modular pipelines. We present BEV-Diffusion, a camera-only model that learns a bird's-eye view (BEV) traversability map, its uncertainty and the path in a single network. A frozen DINOv2 backbone encodes consecutive RGB frames, which inverse perspective mapping and Bayesian fusion lift into the BEV, and a ranking loss keeps outdoor traversability graded rather than binary, separating asphalt from grass (median 0.94 vs. 0.75). On this map, a transformer denoiser generates polar waypoints in a few DDIM steps, and uncertainty-aware guidance refines the path and ends it where the map is no longer reliable. At deployment, a metric BEV memory keeps obstacles after they leave the camera view and warm starts each sample from the previous, ego-motion compensated path. Because the planner jointly learns the path, the map, and its uncertainty, it selects routes through terrain it can confidently traverse. In an unseen courtyard, a wheeled-legged robot autonomously traveled 120\,m, completing 96\% of the route without intervention, compared with 75--88\% for the baselines. Even under low-light conditions after sunset, the proposed model completed the route without requiring any intervention. Retraining only the BEV head, about 3\% of the parameters, adapts the model to a small legged robot in 2.3\,min. Indoors, the memory gets this robot out of maze-like traps: it retains 62--66\% of the nearby out-of-view obstacles, against at most 3\% for the baselines, and the robot achieves 67--100\% task progress, compared with 20--40\% for the evaluated baselines.
\end{abstract}

\begin{IEEEkeywords}
Visual Navigation, Diffusion Models, Bird's-Eye-View Representation,
Traversability Estimation.
\end{IEEEkeywords}

\section{Introduction}
\label{sec:intro}
Legged robots are increasingly deployed in search and rescue, inspection and exploration~\cite{bellicoso2018advances}, where they must navigate unknown terrain without a prior map. This requires estimating robot-specific traversability and generating feasible paths from visual observations. For legged robots, the ability to step over and climb obstacles depends on the body, so that the same terrain can be crossable for one robot and blocking for another. Geometric approaches provide physically grounded terrain information, but they rely on depth sensing or LiDAR and often on separate mapping and planning stages~\cite{mattamala2025wild,cho2026gsat}.   

\begin{figure}[t]
  \centering
  \includegraphics[width=0.85\columnwidth]{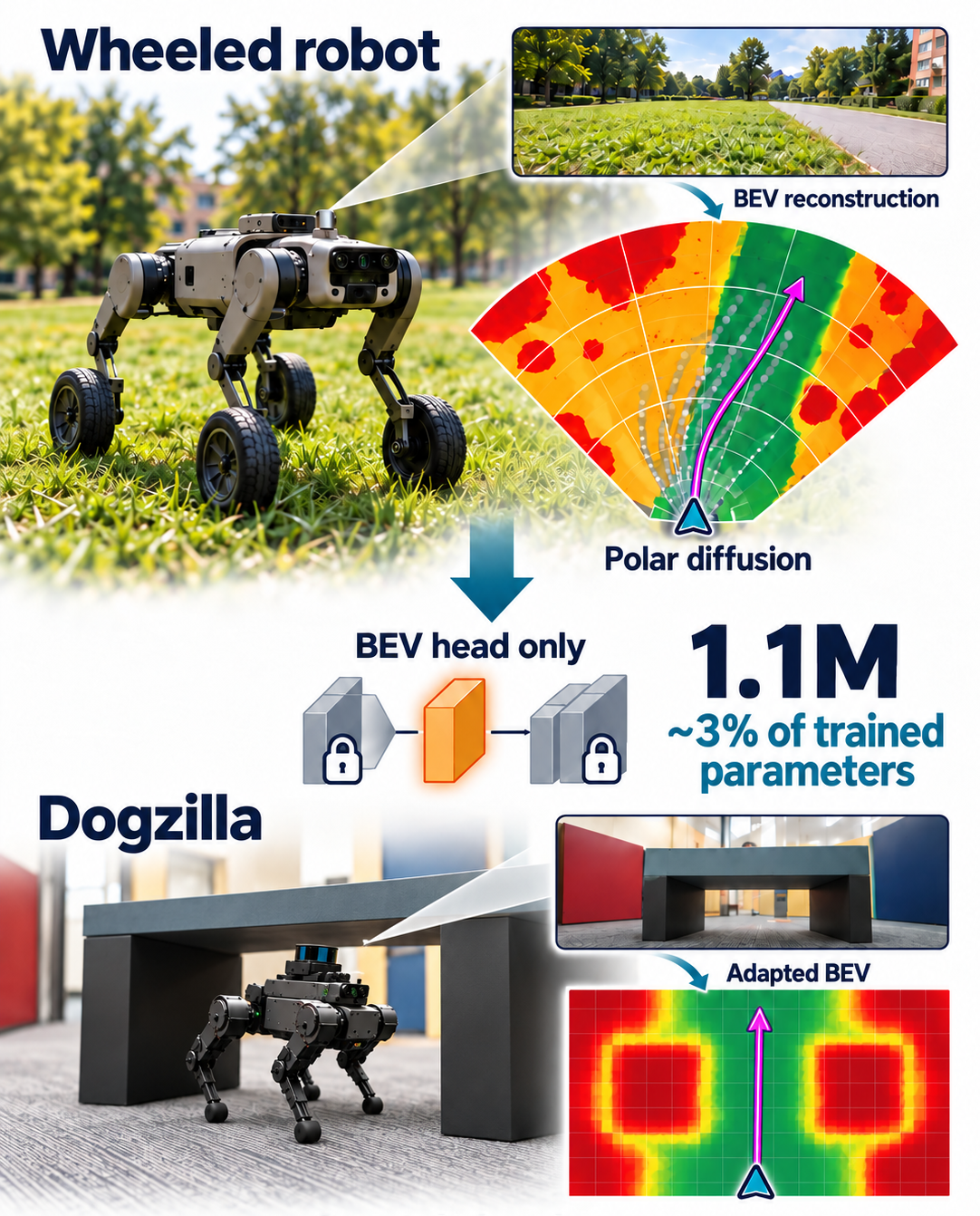}
  \caption{%
  \textbf{BEV-Diffusion on two legged robots.}
  Top: on the wheeled quadruped, the model reconstructs graded BEV
  traversability from the camera and generates the path on it by polar
  diffusion. Bottom: retraining only the BEV head, 1.1M parameters or
  about 3\% of the trained ones, adapts the model in 2.3\,min to the
  small quadruped (Dogzilla S2), which plans through an arch. Images,
  maps and paths are illustrative.%
}
\vspace{-5mm}
  \label{fig:teaser}
\end{figure}

To avoid this, diffusion policies generate paths directly from visual experience. However, they either plan from images without a notion of robot-specific traversability~\cite{sridhar2024nomad}, need depth to recover metric waypoints~\cite{zhura2026embodieddiffusion}, or again rely on a map~\cite{liu2024dipper}. A single camera also perceives less reliably far ahead, in low light and outside its view, yet these policies commit to the whole path without estimating where their perception becomes unreliable. Moreover, since traversability depends on the body, a model trained on one robot does not directly transfer to another. General navigation models reach new robots by training on large multi-robot datasets~\cite{shah2023gnm}, but they do not model traversability, which differs between bodies. These limitations motivate a unified approach that predicts traversability, its uncertainty and metric paths from monocular images alone, commits only where its map is reliable, retains what the robot has already seen, and adapts to a new robot from little data.

We therefore propose BEV-Diffusion, which learns graded BEV traversability, its uncertainty and the path jointly from consecutive RGB frames. Our main contributions are:
\begin{itemize}
     \item \textbf{Joint BEV traversability and trajectory learning:} A camera only framework jointly learns graded traversability, uncertainty and navigation trajectories in a shared BEV representation. The traversability field directly guides trajectory generation, enabling adaptation by updating only the robot specific BEV head while keeping the trajectory generator frozen. This avoids a separate mapping pipeline and projection of image space traversability scores. Ranking supervision preserves terrain preferences, while uncertainty guides cautious path refinement.

    \item \textbf{Memory-assisted navigation:} An ego-motion-compensated BEV memory that retains obstacles outside the current field of view and supports replanning and escape maneuvers. In indoor experiments, it retains 62--66\% of nearby out-of-view obstacles, compared with at most 3\% for the memory-free baselines. In the two trap-like scenes, our models complete 67--100\% of the task, compared with 20--40\% for the evaluated baselines, none of which escapes.
   \item \textbf{Efficient cross-robot adaptation:}
Retraining only the BEV head, about 3\% of the parameters, adapts the model in 2.3\,min from the 41\,kg Agibot D1 Max wheeled quadruped to the much smaller Dogzilla S2 legged robot, while keeping the shared backbone and trajectory generator frozen. Despite changes in robot size, camera viewpoint and environment, from outdoor terrain to indoor mazes, the adapted model achieves 89\% closed loop task progress, compared with 96\% for training from scratch in 19\,min.
\end{itemize}

\section{Related Work}
\label{sec:related}

\subsection{Traversability Estimation}
Traversability tells a robot which terrain it can cross and which it should prefer, and thus which paths are safe. It is usually predicted in the image by a perception module and then projected into a map built from LiDAR or depth point clouds and the robot's pose, on which a separate planner operates, as in WVN~\cite{mattamala2025wild}. Geometric traversability methods estimate terrain suitability from slope, roughness, and clearance~\cite{cho2026gsat}. Learning-based approaches such as WVN and WayFAST~\cite{gasparino2022wayfast} incorporate robot experience but rely on generalization beyond previously traversed regions. Many self-supervised methods, moreover, learn traversability as binary, labelling the ground the robot has walked on as traversable and the rest as not~\cite{ma2025imost}. A binary map only separates free space from obstacles, whereas graded traversability also lets the planner prefer easier terrain, such as asphalt over grass, when several paths are free. Language-guided methods, including AnyTraverse~\cite{sahu2025anytraverse} and EmbodiedDiffusion~\cite{zhura2026embodieddiffusion}, introduce such semantic terrain preferences. BEV-Diffusion instead predicts graded traversability directly in the BEV from the camera, without a point cloud map, combining geometric and semantic supervision, with uncertainty estimates used during path refinement.

\subsection{BEV Reconstruction}
A camera observes the scene in perspective, while a ground robot moves and plans on the ground plane, where distances, free space, and its own footprint are metric. A bird's-eye view (BEV) is a top-down projection of the scene onto this plane: maps from consecutive frames can be fused in it and paths planned on it directly, but lifting image features into it is ill-posed without depth. BEVFormer~\cite{li2024bevformer} learns spatial and temporal BEV features through attention, and SkyEye~\cite{gosala2023skyeye} learns BEV semantic maps from frontal view labels alone through a shared voxel representation. These methods, however, focus on perception rather than trajectory generation. TrajDiff~\cite{gui2025trajdiff} couples the two for autonomous driving: a diffusion transformer generates trajectories from BEV features that are supervised only by a heatmap of the future ego path. Its main model fuses camera and LiDAR at test time, and its BEV encodes where the vehicle drove rather than which terrain is traversable, without adaptation to a new robot. BEV-Diffusion instead predicts a BEV map of robot-specific, graded traversability from frontal RGB features, fuses it over time on top of planar IPM and plans on it. It uses LiDAR only to generate training labels and navigates from the camera alone.

\subsection{Diffusion Based Navigation}

Once traversability is estimated in a spatial representation, the next challenge is generating feasible trajectories from it. A scene often admits several valid paths, for example on either side of an obstacle, and a regressed path can average them into one that crosses the obstacle. Diffusion models~\cite{ho2020denoising,chi2023diffusion} instead learn the full distribution of paths, and their iterative sampling can be guided at test time by a cost such as collision or traversability, without retraining. NoMaD~\cite{sridhar2024nomad} learns goal-masked visual navigation policies, while DiPPeR~\cite{liu2024dipper} and SketchPlan~\cite{norelius2025sketchplan} generate trajectories conditioned on spatial maps or sketches, and NaviDiffusor~\cite{zeng2025navidiffusor} incorporates collision-aware gradient guidance. EmbodiedDiffusion~\cite{zhura2026embodieddiffusion} removes the mapping stage altogether by guiding diffusion with traversability predicted in the image, but it plans in the image plane, so it needs depth to recover metric waypoints and loses obstacles once they leave the camera view. BEV-Diffusion keeps this mapless design but learns traversability and paths jointly in the BEV, so its paths are metric without depth, are refined by the map uncertainty, and a metric BEV memory keeps obstacles after they leave the view.

\section{Method}
\label{sec:method}
\begin{figure*}[!t]
  \centering
  \includegraphics[width=0.85\textwidth]{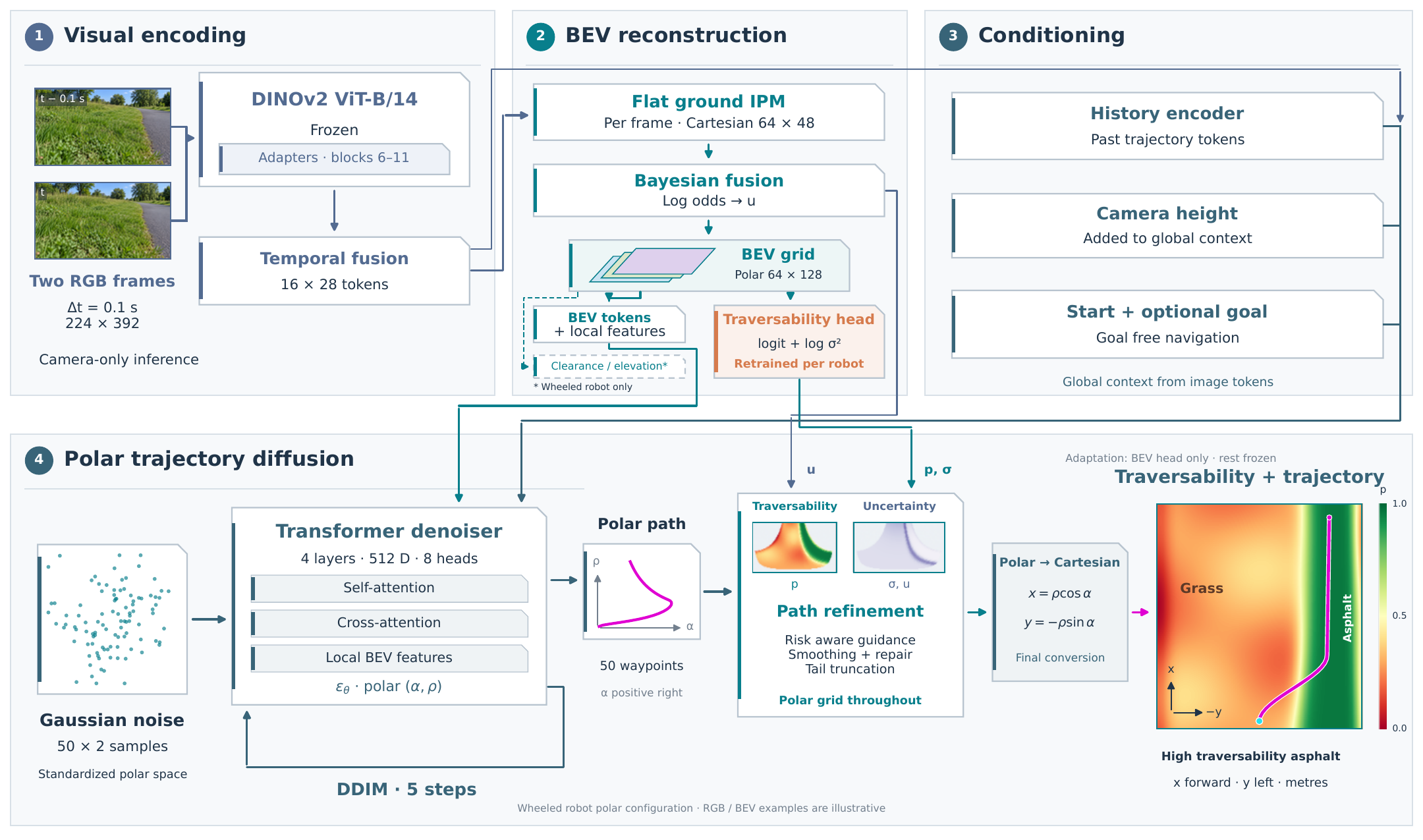}
  \vspace{-2mm}
  \caption{%
  \textbf{Overview of BEV-Diffusion.} A frozen DINOv2 encodes two RGB frames, which are projected and fused into a polar BEV, where the BEV head predicts the traversability $p$ with its epistemic ($u$) and aleatoric ($\sigma$) uncertainty. A transformer denoiser generates polar waypoints on this BEV, which $p$, $u$ and $\sigma$ then refine and truncate. Adaptation retrains only the BEV head. Images and maps
  are illustrative.
  }
  \vspace{-3mm}
  \label{fig:architecture}
\end{figure*}

BEV-Diffusion predicts graded BEV traversability, uncertainty estimates, and a 50-waypoint polar path from two RGB frames separated by 0.1 s (Fig.~\ref{fig:architecture}). Camera height, recent motion history, and an optional goal provide additional conditioning. A pretrained visual encoder and geometric projection construct the BEV representation, which conditions diffusion-based trajectory generation and uncertainty-aware refinement. The perception and trajectory models are trained jointly. 

\subsection{Polar BEV and trajectory space}
\label{sec:bevspace}
The BEV covers $[d_{\min}, d_{\max}] = [0.3, 6.0]$\,m ahead and
$\pm 2.5$\,m to the sides for the wheeled quadruped, and
$[0.35, 2.0]$\,m and $\pm 1.3$\,m for the small quadruped. BEV features
and paths share a polar grid of $N_\rho \times N_\alpha = 64 \times 128$
cells that spans $\pm 90^\circ$; its range resolution matches the
0.09\,m of the Cartesian labels. A waypoint at forward position $f$ and
lateral position $r$ has the normalised polar coordinates
\begin{equation}
  \alpha_n = \tfrac{1}{\pi}\bigl(\operatorname{atan2}(r, f) + \tfrac{\pi}{2}\bigr),
  \quad
  \rho_n = \frac{\sqrt{f^2 + r^2} - d_{\min}}{d_{\max} - d_{\min}} ,
  \label{eq:polar}
\end{equation}
so a path is $\mathbf{x}_0 \in [0,1]^{T_w \times 2}$.

\vspace{-3mm}

\subsection{Visual encoder and conditioning}
\label{sec:encoder}
A frozen DINOv2 ViT-B/14~\cite{oquab2023dinov2} encodes each frame of $224 \times 392$ pixels into $16 \times 28$ patch tokens of dimension $D{=}768$; bottleneck adapters~\cite{houlsby2019parameter} ($d_a{=}64$) in its last six blocks are its only trained parameters. Temporal attention fuses the frames into the tokens $\mathbf{Z}$, whose mean gives a global context vector $\mathbf{c}$, and the tokens $\mathbf{H}^{(t)}$ of each frame feed the BEV. Three signals enter through projections initialised to zero: up to 500 past positions of the robot, encoded by a Transformer; the camera height above the local ground; and the goal, rendered as a Gaussian heatmap in the image. The goal is hidden in half of the training batches, so that the model learns to choose where the path ends.

\subsection{BEV construction with uncertainty}
\label{sec:bev}
Under a flat ground assumption with camera height $h_c$, each cell at $(f, r)$ of an intermediate $64 \times 48$ Cartesian grid samples the patch features at the pixel
\begin{equation}
  u = f_x \tfrac{r}{f} + c_x, \qquad v = f_y \tfrac{h_c}{f} + c_y ,
  \label{eq:ipm}
\end{equation}
after past frames are reprojected with the relative camera pose. Each frame thus gives a traversability logit $\ell^{(t)}_c$ per cell, and we fuse the frames that observe the cell, $v^{(t)}_c \in \{0,1\}$, by adding their log odds:
\begin{equation}
  \ell_c = \textstyle\sum_t v^{(t)}_c \ell^{(t)}_c, \quad
  p_c = \sigma(\ell_c), \quad
  u_c = \mathcal{H}(p_c) \in [0, \log 2].
  \label{eq:bayes}
\end{equation}
The binary entropy $u_c$ is the epistemic uncertainty, largest where the frames disagree; cells that no frame observes receive a learned embedding for unknown space. The fused features are resampled to the polar grid together with $u$, and a convolutional head predicts for every cell the traversability logit $\hat m_c$ and the log variance $\lambda_c$~\cite{kendall2017}, which gives the aleatoric uncertainty
\begin{equation}
  \sigma_c = \exp\bigl(\tfrac{1}{2} \lambda_c\bigr).
  \label{eq:uncertainty}
\end{equation}
For the wheeled quadruped, further heads predict the overhead clearance $\hat h_c \in [0, 2.5]$\,m and the terrain elevation $\hat e_c$. A linear projection turns the BEV into the tokens that the denoiser attends to. Outside the camera view, the map rests on the past frame and on learned scene priors.

\subsection{Trajectory diffusion}
\label{sec:diffusion}
The diffusion runs on standardised coordinates
$\mathbf{z}_0 = (\mathbf{x}_0 - \boldsymbol\mu)/\mathbf{s}$, with the mean and standard deviation of the training paths, so that the waypoint steps are not lost in the noise of the last DDIM steps. A DDPM~\cite{ho2020denoising} with $T_{\mathrm{diff}}{=}1000$ steps and a linear $\beta$ schedule from $10^{-4}$ to $0.02$ adds the noise. The denoiser $\boldsymbol\epsilon_\theta$ is a Transformer with 4 layers, a width of 512 and 8 heads (26M parameters). Each waypoint token combines its noisy coordinates, the BEV feature sampled at these coordinates and a temporal context from the frame tokens. Each layer applies self-attention over the waypoints, cross-attention to the past trajectory tokens and gated cross-attention to the BEV tokens and to the start and goal anchors.

\subsection{BEV memory}
\label{sec:memory}
The predicted map covers only the camera view, so a passed obstacle vanishes from it. At deployment, an ego-centric metric grid of traversability log-odds $L$ keeps it (for the small robot, 2\,cm cells from 1.5\,m behind to 2.5\,m ahead and $\pm 1.5$\,m to the sides). Each step moves the grid by the ego-motion, estimated from the images alone: SIFT features~\cite{lowe2004sift} matched between consecutive frames are projected onto the floor with Eq.~\eqref{eq:ipm}, and a rigid planar motion is fitted with RANSAC~\cite{fischler1981ransac}, which rejects points on walls and furniture. The grid then decays towards unknown, more slowly while the robot stands still, and fuses the new map in the camera view as $L \leftarrow L + \eta\,(1 - u/\log 2)\operatorname{logit} p$, so that uncertain cells count less. The memory does not enter the denoiser. Instead, it warm starts each sample from the previous path, moved by the ego-motion and noised to 35\,\% of the diffusion time, unless the current map contradicts it. It also fills the camera blind zone, keeps the path straight until the body has passed remembered obstacles, and, when the robot makes no progress, turns it in place towards remembered free space.

\subsection{Traversability labels}
\label{sec:supervision}
The traversability $\tau_c \in [0,1]$ of a cell states how easily the robot can cross it, from 0 (impassable) to 1 (free). The LiDAR, the odometry and the IMU serve only to generate these training labels and the camera height; at run time, the camera is the only sensor. For the wheeled quadruped, we fuse 20 LiDAR sweeps before and after each frame with the odometry and fit a local ground plane, which gives every cell an elevation $e_c$ and a slope $\theta_c$. The label is \begin{equation}
  \tau_c = \mathbb{I}_c\, w_c\, \bigl[\beta + (1-\beta)\, s_c\bigr]\,
  r_c\, g_c .
  \label{eq:trav}
\end{equation}
The indicator $\mathbb{I}_c$ requires LiDAR support, a step below 0.80\,m, a slope below $45^\circ$, no obstacle at body height and enough headroom. The geometric factor $g_c$ falls linearly with the step height (1 at 0.05\,m, 0.6 at 0.25\,m, 0.15 at 0.80\,m) and with the slope (1 at $15^\circ$, 0 at $45^\circ$). The semantic weight $w_c$ comes from CLIPSeg~\cite{luddecke2022clipseg}, e.g.\ 1.0 for asphalt and 0.7 for grass. The stability $s_c \in [0.15, 1]$ falls with the vertical acceleration and the tilt that the IMU measured where the robot drove, and $\beta = 0.4$ limits its influence. The roughness $r_c \in [0.35, 1]$ falls with the spread of the LiDAR points around the surface. Only cells with LiDAR returns are labelled; they form the observation mask $\mathcal{O}$.

The small quadruped shakes strongly while it walks, so its IMU signal is dominated by the gait and many images are blurred. It therefore uses the LiDAR labels without the IMU and semantic factors, and blurred frames are removed. On its flat indoor floor, Eq.~\eqref{eq:trav} reduces to $\tau_c = \mathbb{I}_c$: a cell is free up to the first obstacle along each camera ray and blocked behind it. For both robots, the path target is the recorded odometry path, resampled to $T_w$ waypoints.

\subsection{Training objective}
\label{sec:objective}
Map and planner are trained jointly with
\begin{equation}
\begin{aligned}
  \mathcal{L} ={}& \mathcal{L}_{\mathrm{diff}}
  + 0.5\,(\mathcal{L}_{\mathrm{bev}} + \mathcal{L}_{\mathrm{frame}})
  + 0.1\,\mathcal{L}_{\sigma} \\
  &+ 0.3\,(\mathcal{L}_{\mathrm{path}} + \mathcal{L}_{\mathrm{unc}})
  + 0.05\,\mathcal{L}_{\mathrm{pred}} \\
  &+ \mathcal{L}_{\mathrm{clear}}
  + 0.5\,(\mathcal{L}_{\mathrm{elev}} + \mathcal{L}_{\mathrm{start}}
  + \mathcal{L}_{\mathrm{goal}}).
\end{aligned}
\label{eq:total}
\end{equation}
The diffusion term is the noise prediction loss $\mathcal{L}_{\mathrm{diff}} = \|\boldsymbol\epsilon_\theta(\mathbf{z}_t, t, \mathcal{C}) - \boldsymbol\epsilon\|^2$ with the conditioning $\mathcal{C}$. The map term is an attenuated cross entropy~\cite{kendall2017} on the observed cells plus a ranking term,
\begin{equation}
  \mathcal{L}_{\mathrm{bev}} =
  \frac{\sum_{c \in \mathcal{O}} \kappa_c\,
        \mathrm{BCE}(\hat m_c + \sigma_c \epsilon,\, \tau_c)}
       {\sum_{c \in \mathcal{O}} \kappa_c}
  + 0.5\,\mathcal{L}_{\mathrm{rank}} ,
  \label{eq:lbev}
\end{equation}
with $\epsilon \sim \mathcal{N}(0,1)$ averaged over 5 samples. Unobserved cells are excluded, because supervising them as blocked would teach the model that distant space is blocked. Only about 7\,\% of flat terrain is blocked, so the weight $\kappa_c = 1 + 3\,\mathrm{clip}\bigl((0.3 - \tau_c)/0.3,\,0,\,1\bigr)$ raises the influence of blocked cells. The ranking term preserves graded levels such as asphalt and grass: for random pairs of cells whose labels differ by more than 0.15, it penalises predicted differences that are too small,
\begin{equation}
  \mathcal{L}_{\mathrm{rank}} = \operatorname{mean}
  \bigl[\,|\Delta\tau| - \operatorname{sign}(\Delta\tau)\,\Delta\hat p\,\bigr]_+ .
  \label{eq:rank}
\end{equation}
$\mathcal{L}_{\mathrm{frame}}$ applies the cross entropy to the logits
$\ell^{(t)}_c$ of each frame before the fusion, and
$\mathcal{L}_\sigma = \operatorname{mean}_c (\lambda_c - \lambda^\ast_c)^2$
regresses the log variance to a target that is high for ambiguous or unlabelled cells, $\lambda^\ast_c = \log 4 - o_c\,|2\tau_c - 1|\log 80$, where $o_c$ marks labelled cells. Three terms couple the maps to the path by sampling them at the waypoints:
\begin{equation}
\begin{aligned}
  \mathcal{L}_{\mathrm{path}} &= \operatorname{mean}_k\,
  \mathrm{BCE}\bigl(\hat m(\mathbf{x}^\ast_k),\, 1\bigr), \\
  \mathcal{L}_{\mathrm{unc}} &= \operatorname{mean}_k\,
  u(\mathbf{x}^\ast_k) / \log 2, \\
  \mathcal{L}_{\mathrm{pred}} &= (1 - t/T_{\mathrm{diff}})^2
  \operatorname{mean}_k\, u(\hat{\mathbf{x}}_k) / \log 2 ,
\end{aligned}
\label{eq:lpath}
\end{equation}
where $\mathbf{x}^\ast$ is the recorded path and $\hat{\mathbf{x}}$ the clean path that the model predicts from $\mathbf{z}_t$; the last term makes the model avoid uncertain cells already during denoising. The clearance and elevation heads use a smooth $\ell_1$ loss on $\mathcal{O}$, and two linear heads on $\mathbf{c}$ regress the start and the goal of the path. For the small quadruped, we omit the clearance and elevation terms, the weight $\kappa_c$ and $\mathcal{L}_{\mathrm{rank}}$. We train for 30 epochs with AdamW, a learning rate of $10^{-4}$ with cosine decay and an effective batch size of 32.

\subsection{Adaptation to a new robot}
\label{sec:adaptation}
To transfer the model to a new robot, we keep all weights of the source model and retrain only the traversability head, 1.1M of the 40M trained parameters, on the labels of the target robot with $\mathcal{L}_{\mathrm{bev}}$, $\mathcal{L}_\sigma$ and $\mathcal{L}_{\mathrm{path}}$. The geometry is recomputed for the target robot, including the camera model, the projection of Eq.~\eqref{eq:ipm} and the BEV extent, while the frozen denoiser keeps the path statistics $(\boldsymbol\mu, \mathbf{s})$ of the source robot.

\subsection{Inference}
\label{sec:inference}

We generate goal-free paths using five deterministic DDIM steps~\cite{song2020denoising}. Predicted waypoints are refined through traversability- and uncertainty-based gradient guidance, followed by corridor snapping using the risk-adjusted score ($p - 0.5\,\sigma$). The path is smoothed and repaired to maintain a minimum obstacle margin of half the robot width. Waypoints are truncated when their confidence $p\,(1 - 0.7\,u/\log 2) - 0.15\,\min(\sigma, 2)$ falls below 0.5. The complete inference pipeline requires 42\,ms per plan in bfloat16 on an RTX 5090, supporting operation at 20\,Hz. 
\section{Experiments}
\label{sec:experiments}

\subsection{Robots and datasets}
\label{sec:datasets}
We evaluate the method on two quadrupeds that differ strongly in size, camera and environment.

\textbf{Wheeled quadruped.}
The Agibot D1 Max (Inchbot ZSM-1) is a 41\,kg quadruped with wheels on its feet ($0.93 \times 0.48 \times 0.585$\,m). Its front camera ($1920 \times 1080$, undistorted to $960 \times 540$) sits about 0.60\,m above the ground. A forward-facing RS-AIRY LiDAR, odometry and an IMU provide the training and evaluation labels only; navigation uses the camera alone.

We recorded four flat outdoor scenes (asphalt, grass and two pebble areas), 4{,}841 frames in total. A contiguous 15\,\% segment inside each scene is held out for validation, giving 3{,}921 training and 688 validation windows.

\textbf{Small quadruped.} The Dogzilla S2 is a small legged robot. Its fisheye camera ($1920 \times 1080$, rectified to $960 \times 540$) is only 0.162\,m above the floor. We recorded four runs with 4{,}740 frames in an indoor room with mats, foam blocks, arches, a table on thin chrome legs, poles on round bases and boxes. One complete run (462 windows) is held out for validation, and the other three runs (1{,}782 windows) are used for training. Compared with the wheeled robot, the camera is almost four times lower, the lens and the scene type differ, and the BEV covers 2\,m instead of 6\,m.

\subsection{Evaluation protocol and metrics}
\label{sec:metrics}
Each configuration is trained three times; we report the mean, and in the ablation also the standard deviation. All runs use the same validation windows and diffusion noise, and no planner goal (Sec.~\ref{sec:inference}).

\textit{BEV reconstruction.} As in semantic BEV
segmentation~\cite{gosala2023skyeye}, we score all labelled cells of the BEV, including those outside the camera view. Since 96\,\% of them are traversable, a binary IoU is uninformative, so we report the level mIoU over blocked ($\tau < 0.3$), rough ($< 0.6$), moderate ($< 0.85$) and easy cells. The blocked recall (Blk.\ rec.) is the share of blocked cells ($\tau \le 0.15$) predicted as blocked ($\hat p_c < 0.5$), on the whole BEV and outside the view (Out); we also report the expected calibration error of $\hat p_c$.

\textit{Paths.} Trav.\ is the mean label along a path, Easy the share of its length on easy terrain, Clear.\ the median minimum distance to a blocked cell, and Cross.\ the share of paths that enter one. As shorter paths keep more distance, we mark variants whose paths are much shorter than the recorded ones, which we score as a reference. 
\textit{Real-world metrics.} LiDAR scans, labelled as in training, serve as ground truth. Besides the level mIoU and blocked recall in the camera view, we report the share of LiDAR obstacles within 1\,m outside the view that the map marks as blocked (out-of-view obst.\ known) and the range error (Range err.), the median distance along camera rays between the first predicted and the first LiDAR obstacle. Path ok is the share of paths whose followed part keeps half the robot width from every LiDAR obstacle (8\,cm over 0.1--0.3\,m for the small robot, 20\,cm over the first metre for the wheeled one), Clear.\ the median of this distance, and $<$5\,cm time the share of scans in which the body is closer than 5\,cm to an obstacle.

\subsection{Ablation study}
\label{sec:ablation}
\begin{table*}[t]
\centering
\vspace{2mm}
\caption{Ablation on the wheeled quadruped, mean $\pm$ std over three training runs. BEV metrics on the whole BEV; Out: outside the camera view. Paths without a planner goal. Image projection: image-space traversability projected onto a flat ground plane. The guidance row reuses the checkpoints of the full model. Bold: best. $^\ddagger$Paths about half as long as the recorded ones, excluded from the comparison of the path metrics.}
\label{tab:ablation}
\setlength{\tabcolsep}{5pt}
\begin{tabular}{lcccccc}
\toprule
 & \multicolumn{3}{c}{BEV reconstruction} & \multicolumn{3}{c}{Paths} \\
\cmidrule(lr){2-4}\cmidrule(lr){5-7}
Variant & Level mIoU $\uparrow$ & Blk.\ rec.\ $\uparrow$ & Blk.\ rec.\ Out $\uparrow$
 & Trav.\ $\uparrow$ & Easy [\%] $\uparrow$ & Clear.\ [m] $\uparrow$ \\
\midrule
Image projection, no learned BEV & $0.198$ & $0.056$ & $0.000$ & -- & -- & -- \\
\midrule
Full model (polar BEV, $T{=}2$) & $0.319\pm0.007$ & $0.695\pm0.013$ & $0.755\pm0.011$ & $0.831\pm0.007$ & $45.7\pm2.7$ & $1.14\pm0.04$ \\
Cartesian BEV & $0.320\pm0.001$ & $0.624\pm0.006$ & $0.693\pm0.037$ & $0.830\pm0.008$ & $45.2\pm3.8$ & $1.15\pm0.04$ \\
w/o uncertainty & $0.327\pm0.004$ & $0.664\pm0.023$ & $0.751\pm0.029$ & $\mathbf{0.850\pm0.022}$ & $\mathbf{54.7\pm7.5}$ & $0.89\pm0.13$ \\
$T{=}1$ frame$^\ddagger$ & $\mathbf{0.328\pm0.003}$ & $0.679\pm0.012$ & $0.746\pm0.044$ & $0.820\pm0.007$ & $41.2\pm3.2$ & $1.25\pm0.04$ \\
$T{=}3$ frames & $0.320\pm0.007$ & $\mathbf{0.700\pm0.029}$ & $\mathbf{0.808\pm0.016}$ & $0.837\pm0.009$ & $48.5\pm2.8$ & $1.04\pm0.06$ \\
w/o ranking loss & $0.304\pm0.003$ & $0.648\pm0.014$ & $0.738\pm0.011$ & $0.826\pm0.003$ & $43.7\pm2.0$ & $1.15\pm0.02$ \\
w/o traversability guidance & \multicolumn{3}{c}{\textit{same map as the full model}} & $0.793\pm0.005$ & $32.9\pm1.3$ & $\mathbf{1.51\pm0.03}$ \\
\midrule
\textit{Recorded path} & -- & -- & -- & $0.779$ & $30.2$ & $1.60$ \\
\bottomrule
\vspace{-5mm}
\end{tabular}
\end{table*}

\subsubsection{BEV representation} Table~\ref{tab:ablation} shows that
learned BEV reconstruction substantially improves blocked-cell recall over direct image projection (0.695 vs.\ 0.056). Polar and Cartesian representations achieve similar mIoU and path traversability, but the polar representation detects more blocked cells (0.695 vs.\ 0.624), at an additional cost of 2.6\,ms per plan. The ranking loss keeps the map graded rather than binary: without it, the level mIoU drops from 0.319 to 0.304, and the share of correctly assigned rough cells from 41\,\% to 25\,\%.

\subsubsection{Uncertainty and temporal context} Uncertainty-aware planning increases obstacle clearance from 0.89 to 1.14\,m by truncating trajectories in ambiguous regions. Additional frames improve out-of-view blocked-cell recall (0.746 with one frame vs.\ 0.808 with three), but correlated observations raise the calibration error from 0.043 with two frames to 0.051 with four. We therefore use two frames. 

\subsubsection{Trajectory guidance.} Traversability guidance increases mean path traversability from 0.793 to 0.831, exceeding the recorded trajectories (0.779), while increasing the proportion of path length on easy terrain from 32.9\,\% to 45.7\,\%. This improvement comes at the cost of reduced obstacle clearance, reflecting a tradeoff between terrain preference and conservative planning.

\subsection{Real-world experiments}
\label{sec:real_world}
We evaluate closed-loop outdoor exploration with the wheeled-legged robot and indoor exploration with the small legged robot, both against the baselines. In all experiments, the robots navigate without a prior map and without localisation: the model plans from the camera stream alone, and the BEV memory of the small robot estimates its ego-motion from the floor in the images.

\label{sec:real_world}
\subsubsection{Baselines.} For the modular pipeline, we adapt BEVFormer~\cite{li2024bevformer} to a single front camera. To exclude the backbone as a source of difference, it uses our frozen DINOv2 ViT-B/14~\cite{oquab2023dinov2} with a simple feature pyramid~\cite{li2022vitdet}, receives the same two frames and is trained with our labels and map losses (Sec.~\ref{sec:objective}). We combine its map with two planners. The first is A*~\cite{hart1968formal} on a cost map, in which cells with a risk adjusted traversability $\hat p - 0.5\,\sigma$ below 0.3 are obstacles inflated by half the robot width. The second is a map conditioned diffusion planner in the spirit of DiPPeR~\cite{liu2024dipper}, which shares our path representation, sampler and corridor snap. As in a modular system, it is trained on the labelled maps and receives the predicted maps at test time. As the end-to-end baseline, we use TG-Diff (Traversability-Guided Diffusion), a diffusion policy whose sampling is guided by a CLIPSeg traversability map instead of the RGB image, adapted from DroneDiffusion~\cite{valuev2026dronediffusionuncertaintyguideddiffusionpath} and retrained on our robots. All methods are trained three times and evaluated with the protocol of Sec.~\ref{sec:metrics}.

\subsubsection{Outdoor exploration with the wheeled-legged robot}
\label{sec:rw_wheeled}
We deployed BEV-Diffusion without persistent BEV memory in an unseen courtyard containing grass, posts, bins, and parked bicycles. The robot traveled 120\,m autonomously across six goal-free sessions, including four after sunset (\emph{Scene~1}) and two in daylight (\emph{Scene~2}). \emph{Scene~3} adds more diverse terrains: grass and asphalt. For comparison, all baselines processed the same recorded camera streams, with LiDAR-derived labels used exclusively for evaluation. 

\begin{table*}[t]
  \centering
  \vspace{2mm}
  \caption{Wheeled quadruped outdoors. Scenes~1 and~2: unseen courtyard after
 sunset and in daylight; Scene~3: open validation area of the training scenes (asphalt, grass).}
  \label{tab:wheeled}
  \small
  \setlength{\tabcolsep}{3.5pt}
  \begin{tabular}{llccccccccc}
\toprule
Scene & Method & mIoU & Blk. rec. & Easy & Cross. & Cross. full & Len. m & Interv. & Succ. & Train min \\
\midrule
\multirow{4}{*}{Scene 1} & Ours & \textbf{0.19} & 40\% & 84\% & \textbf{13\%} & \textbf{18\%} & 2.8 & \textbf{0} & \textbf{100\%} & \textbf{21.3} \\
 & BEVFormer + A$^*$ & \textbf{0.19} & \textbf{69\%} & \textbf{86\%} & 18\% & 30\% & 5.6 & 10 & 81\% & 38.8 \\
 & BEVFormer + Diff. & \textbf{0.19} & \textbf{69\%} & \textbf{86\%} & 27\% & 46\% & 6.6 & 6 & 81\% & 41.0 \\
 & TG-Diff & -- & -- & 81\% & 17\% & 43\% & 5.1 & 5 & 90\% & 40.0 \\
\midrule
\multirow{4}{*}{Scene 2} & Ours & \textbf{0.18} & 37\% & \textbf{93\%} & 14\% & \textbf{16\%} & 1.5 & \textbf{1} & \textbf{67\%} & \textbf{21.3} \\
 & BEVFormer + A$^*$ & \textbf{0.18} & \textbf{55\%} & 82\% & 25\% & 71\% & 5.5 & 4 & 33\% & 38.8 \\
 & BEVFormer + Diff. & \textbf{0.18} & \textbf{55\%} & 86\% & 26\% & 88\% & 7.1 & 3 & \textbf{67\%} & 41.0 \\
 & TG-Diff & -- & -- & \textbf{93\%} & \textbf{3\%} & 64\% & 4.9 & \textbf{1} & \textbf{67\%} & 40.0 \\
\midrule
\multirow{4}{*}{Scene 3} & Ours & 0.33 & \textbf{71\%} & 65\% & \textbf{0\%} & \textbf{0\%} & 5.3 & \textbf{0} & \textbf{100\%} & \textbf{21.3} \\
 & BEVFormer + A$^*$ & \textbf{0.38} & 55\% & 78\% & \textbf{0\%} & \textbf{0\%} & 6.0 & \textbf{0} & \textbf{100\%} & 38.8 \\
 & BEVFormer + Diff. & \textbf{0.38} & 55\% & \textbf{93\%} & \textbf{0\%} & \textbf{0\%} & 5.7 & \textbf{0} & \textbf{100\%} & 41.0 \\
 & TG-Diff & -- & -- & 53\% & \textbf{0\%} & \textbf{0\%} & 6.0 & \textbf{0} & \textbf{100\%} & 40.0 \\
\bottomrule
\end{tabular}

\end{table*}
Table~\ref{tab:wheeled} summarises the results. We count an intervention when a method's path runs into a LiDAR obstacle at least 10\,cm high within the first meter in view and is not re-planned away within 0.5\,s, and report the share of the driven route (in 5\,m stretches) without one. In Scenes~1 and~2, our model succeeds on 96\,\% of the route, against 75--88\,\% for the baselines, with no failure after sunset; its one failure coincides with the operator stopping the robot at a bush.

\begin{figure}[t]
  \centering
  \makebox[\columnwidth][l]{\footnotesize\textbf{Scene 1} \quad after sunset: the baselines head into a door wall 1.6\,m ahead}\\[1pt]
  \includegraphics[width=\columnwidth]{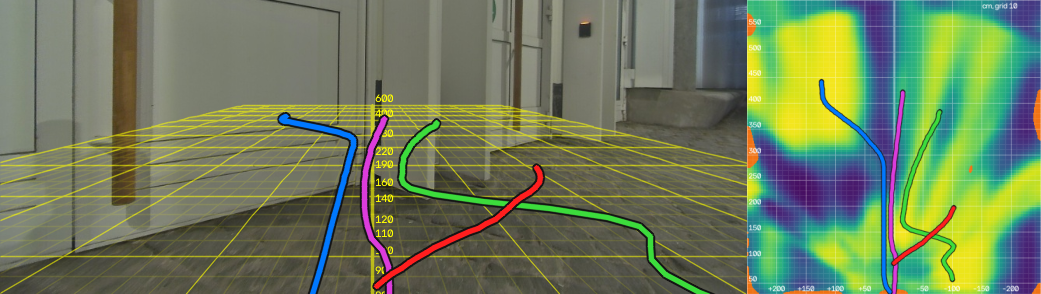}\\[4pt]
  \makebox[\columnwidth][l]{\footnotesize\textbf{Scene 2} \quad daylight: ours ends at 1.6\,m, the baselines cross obstacles}\\[1pt]
  \includegraphics[width=\columnwidth]{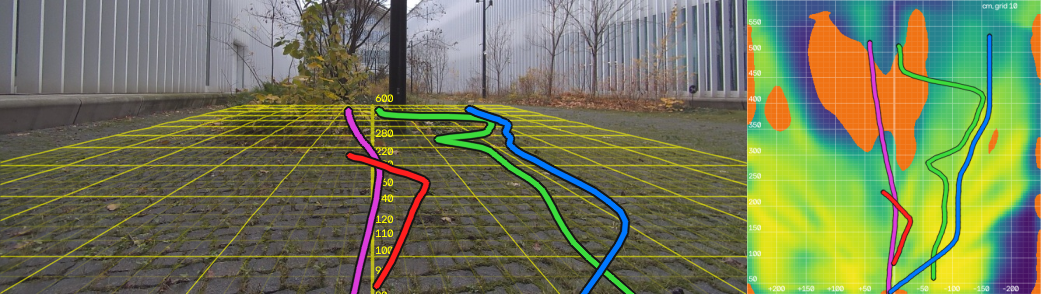}\\[4pt]
  \makebox[\columnwidth][l]{\footnotesize\textbf{Scene 3} \quad grass fills the view: BEV methods move onto the asphalt}\\[1pt]
  \includegraphics[width=\columnwidth]{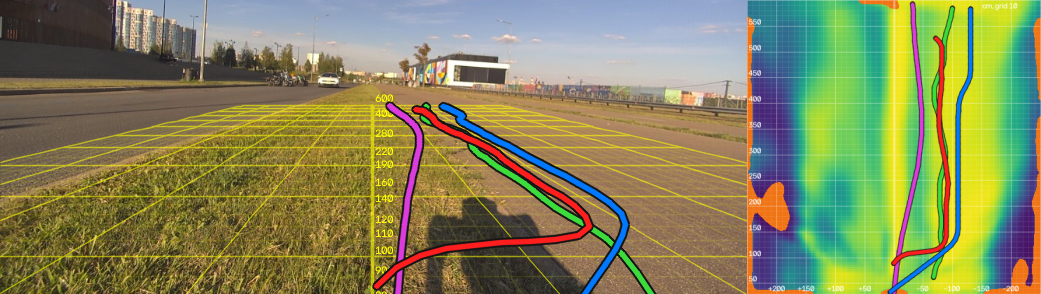}\\[3pt]
  \includegraphics[width=\columnwidth]{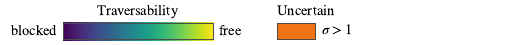}
  \caption{Wheeled quadruped outdoors: camera (left) and our BEV map
    (right, 0.3--6\,m) with the paths of
    \textcolor[RGB]{255,30,30}{\textbf{ours}},
    \textcolor[RGB]{0,120,255}{\textbf{A*}},
    \textcolor[RGB]{40,170,40}{\textbf{diffusion planner}} and
    \textcolor[RGB]{200,50,200}{\textbf{TG-Diff}}. Orange: aleatoric
    uncertainty $\sigma > 1$. In Scene~1, all baseline paths enter the
    LiDAR-labelled door wall within their first 2\,m, while ours turns
    towards the open passage. In Scene~2, only our path avoids both the
    uncertain areas and the LiDAR obstacles further ahead. In Scene~3,
    the BEV methods steer onto the asphalt, TG-Diff stays on the grass.}
  \label{fig:wheeled}
  \vspace{-2mm}
\end{figure}

The difference comes from how far the methods commit. Our paths end where the map becomes uncertain, after 1.5--2.8\,m, whereas the baselines plan 5--7\,m ahead, 42--100\,\% of them through areas that our map marks as highly uncertain ($\sigma > 1$, orange in Fig.~\ref{fig:wheeled}), against 14\,\% of ours. Our model therefore stops more often (17 vs.\ 0--2\,\% of the frames with free ground ahead in Scene~1), but at about 20\,Hz a stopped plan costs only about 1\,cm of travel.

In Scene~3, both BEV maps grade the terrain rather than splitting it into free and blocked (our median 0.94 on asphalt, 0.75 on grass, 0.32 on blocked cells), whereas TG-Diff, relying on CLIPSeg alone, keeps the least of its path on easy terrain (53 vs.\ 65--93\,\%). Building the BEV geometrically (Eq.~\ref{eq:ipm}) also halves training time (21.3
vs.\ 38.8--41.0\,min). Committing only to the reliable part of the map matters most where perception is hardest: after sunset, our model is the only one without an intervention.

\subsubsection{Indoor exploration with the small legged robot}
\label{sec:closed_loop}

\begin{figure}[t]
  \centering
  \makebox[\columnwidth][l]{\footnotesize\textbf{Scene 1} \quad A* finds no path}\\[1pt]
  \includegraphics[width=\columnwidth]{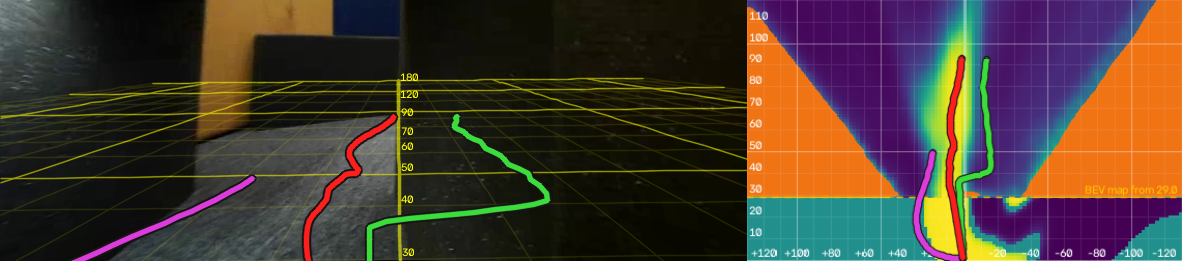}\\[4pt]
  \makebox[\columnwidth][l]{\footnotesize\textbf{Scene 2} \quad ours turns in place}\\[1pt]
  \includegraphics[width=\columnwidth]{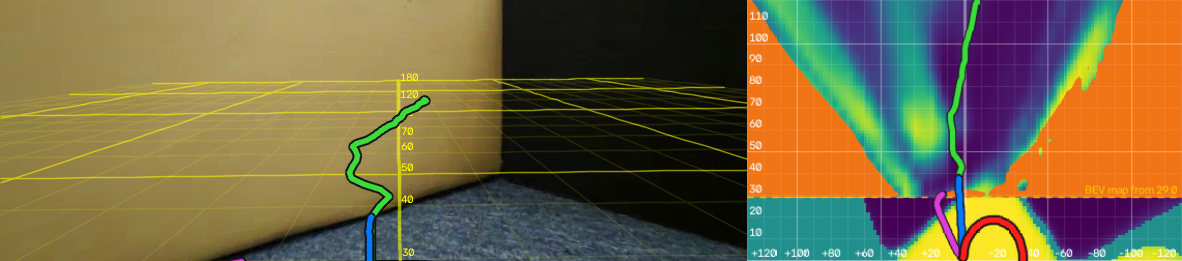}\\[4pt]
  \makebox[\columnwidth][l]{\footnotesize\textbf{Scene 3} \quad baselines graze the arch leg or head into walls}\\[1pt]
  \includegraphics[width=\columnwidth]{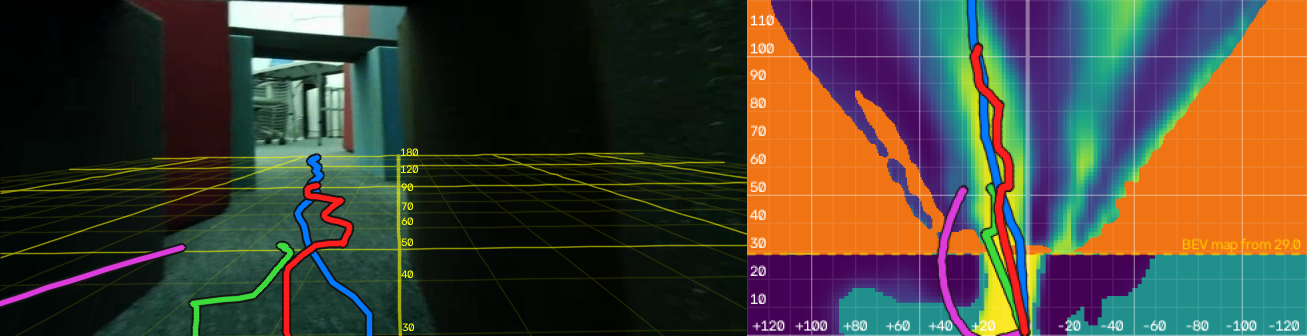}\\[3pt]
  \includegraphics[width=\columnwidth]{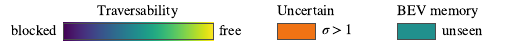}
  \caption{Camera (left) and our BEV map (right) with the paths of
    \textcolor[RGB]{255,30,30}{\textbf{ours}},
    \textcolor[RGB]{0,120,255}{\textbf{A*}},
    \textcolor[RGB]{40,170,40}{\textbf{diffusion planner}} and
    \textcolor[RGB]{200,50,200}{\textbf{TG-Diff}}. Orange: aleatoric
    uncertainty $\sigma > 1$, mostly outside the camera view. Below the
    dashed line lies the camera's blind zone (closer than 29\,cm), which
    the camera cannot see; there the map shows the BEV memory, teal where
    the memory has not seen the floor. Only ours keeps clear of the obstacles; in Scene~2
    it turns in place towards free space kept in the memory.}
  \vspace{-5mm}  
  \label{fig:closed_loop}
\end{figure}

We evaluate closed-loop exploration in three indoor scenarios (Fig.~\ref{fig:closed_loop}): a hallway containing an arch and obstacles (Scene 1), a circular room with a narrow exit and a 60 s time limit (Scene 2), and a corridor with two passable arches followed by an impassable low arch (Scene 3). BEV-Diffusion uses the metric memory described in Sec.~\ref{sec:memory}, while the baselines plan on the current map
only, as memory and escape are part of our contribution. Ground truth is a 2D LiDAR 20.6\,cm above the floor, with the metrics of Sec.~\ref{sec:metrics}. To isolate the memory, we also pass every method's recorded maps offline through it (Table~\ref{tab:closed_loop}, same mem.): the BEVFormer maps then keep 37--72\,\% instead of 0--3\,\% of the out-of-view obstacles, on par with
ours in Scenes~1 and~3, but our map still keeps the most in the round room of Scene~2 (58--59 vs.\ 37--45\,\%), as it marks more of the nearby obstacles in view.

\textbf{Task progress.} We use the first three attempts per method and scene and report the share of the task completed without a collision: the milestones reached in Scene~1 (arch, low block, pole, end of the hallway) and Scene~3 (two arches, turn at the low arch, two arches back), and in Scene~2 the share of the 60\,s the robot keeps moving before it gets stuck, leaving the room counting as 100\,\%.
\begin{table*}[t]
  \centering
  \vspace{2mm}
  \caption{Closed-loop indoor exploration with the small legged robot (Scene~1: hallway, 2: round room, 3: arches).
    Map and path metrics against the onboard LiDAR.}
  \label{tab:closed_loop}
  \small
  \setlength{\tabcolsep}{4pt}
  \begin{tabular}{llccccccccc}
\toprule
Scene & Method & Progress$\uparrow$ & mIoU$\uparrow$ & \multicolumn{2}{c}{Out-of-view obst.$\uparrow$} & Range err.$\downarrow$ & Path ok$\uparrow$ & Clear.$\uparrow$ & $<$5\,cm$\downarrow$ & Train$\downarrow$ \\
\cmidrule(lr){5-6}
 & & [\%] & & live & same mem. & [cm] & ($\geq$8\,cm) & [cm] & time & [min] \\
\midrule
\multirow{5}{*}{Scene 1} & Ours (scratch) & 92\% & 0.31 & \textbf{68\%} & 71\% & \textbf{9.5} & \textbf{82\%} & 11.4 & 28\% & 19.1 \\
 & Ours (adapted) & \textbf{100\%} & \textbf{0.36} & 64\% & 68\% & 10.3 & 78\% & \textbf{12.0} & 20\% & \textbf{2.3} \\
 & BEVFormer + A* & \textbf{100\%} & 0.29 & 1\% & 69\% & 16.5 & \textbf{82\%} & \textbf{12.0} & 14\% & 32.5 \\
 & BEVFormer + Diff. & \textbf{100\%} & 0.26 & 1\% & \textbf{72\%} & 19.0 & 61\% & 9.5 & 29\% & 34.9 \\
 & TG-Diff & 0\% & -- & -- & -- & -- & 42\% & 6.4 & \textbf{0\%} & 40.0 \\
\midrule
\multirow{5}{*}{Scene 2} & Ours (scratch) & 95\% & 0.55 & \textbf{63\%} & 58\% & \textbf{5.8} & 82\% & 16.8 & 9\% & 19.1 \\
 & Ours (adapted) & \textbf{100\%} & 0.46 & 62\% & \textbf{59\%} & 9.2 & 86\% & 18.6 & 7\% & \textbf{2.3} \\
 & BEVFormer + A* & 20\% & \textbf{0.57} & 2\% & 37\% & 9.9 & 74\% & 18.1 & \textbf{0\%} & 32.5 \\
 & BEVFormer + Diff. & 22\% & \textbf{0.57} & 3\% & 45\% & 10.8 & 72\% & 22.2 & 16\% & 34.9 \\
 & TG-Diff & 28\% & -- & -- & -- & -- & \textbf{100\%} & \textbf{25.2} & \textbf{0\%} & 40.0 \\
\midrule
\multirow{4}{*}{Scene 3} & Ours (scratch) & \textbf{100\%} & \textbf{0.35} & \textbf{72\%} & \textbf{68\%} & \textbf{7.9} & 74\% & 12.0 & 33\% & 19.1 \\
 & Ours (adapted) & 67\% & 0.33 & 66\% & 67\% & 11.9 & \textbf{80\%} & \textbf{12.6} & 24\% & \textbf{2.3} \\
 & BEVFormer + A* & 40\% & 0.28 & 0\% & \textbf{68\%} & 8.4 & 65\% & 10.7 & \textbf{20\%} & 32.5 \\
 & BEVFormer + Diff. & 40\% & 0.27 & 1\% & 63\% & 9.3 & 57\% & 9.3 & 26\% & 34.9 \\
\bottomrule
\end{tabular}
\end{table*}

Table~\ref{tab:closed_loop} summarizes the indoor results. In Scene~1, all BEV-based methods reach 92--100\,\% progress, whereas TG-Diff fails to pass the arch. In Scene~2, the BEVFormer baselines achieve comparable
or higher in-view map accuracy but retain only 2--3\,\% of out-of-view obstacles, compared with 62--63\,\% for BEV-Diffusion; our models turn towards remembered free space and keep moving (95--100\,\%), and the adapted one leaves the room in 2 of 3 runs, while the baselines get stuck (20--28\,\%). In Scene~3, only our models turn back at the low arch (67--100 vs.\ 40\,\%). The adapted model achieves task progress comparable to training from scratch (89 vs.\ 96\,\%) after 2.3\,min instead of
19.1\,min of training.

\section{Conclusion}
\label{sec:conclusion}
We presented BEV-Diffusion, a camera-only navigation model that learns a BEV traversability map, its uncertainty and the path in a single network. LiDAR and IMU are needed only for the training labels; at deployment the robot navigates from the camera alone, without a prior map or localisation.

Three insights follow. First, traversability should be learned in the BEV and graded: the learned map detects 69.5\,\% of the obstacles against 6\,\% for an image projection, and keeps asphalt and grass apart (median 0.94 vs.\ 0.75). Second, learning the path together with the map and its uncertainty lets the planner commit only to reliable ground: on the 120\,m our model drove outdoors, 96\,\% of the route passed without an intervention, against 75--88\,\% for the baselines, with none after sunset. Third, a separate robot-specific BEV head makes adaptation cheap: retraining 3\,\% of the parameters for 2.3\,min nearly matched training from scratch (89 vs.\ 96\,\% progress). On the small robot, the BEV memory, which also works with other BEV maps, raised the progress in trap-like scenes to 67--100\,\%, against 20--40\,\%. BEV-Diffusion thus shows that a single camera, a learned BEV and a diffusion planner trained together can replace LiDAR, mapping and localisation at deployment on low-cost legged robots.

\textit{Limitations.} The flat ground projection restricts the model to mostly flat terrain. In an unfamiliar scene, our map detects fewer obstacles than in the training scenes. The Bayesian fusion treats consecutive frames as independent, which limits the number of frames before the map loses calibration. 

\textit{Future work.} Since only the BEV head is robot specific, one model could serve a fleet of legged robots, each adapted in minutes from its own data, and the head could be updated online from the robot's own experience. For uneven terrain such as slopes and stairs, we will condition the model on the camera inclination from the IMU and test the step height and slope terms of Eq.~\eqref{eq:trav}, with a projection that accounts for the predicted elevation.

\balance
\bibliographystyle{ieeetr}
\bibliography{references}

\end{document}